\documentclass[11pt]{article}

\usepackage{acl}

\usepackage{,}
\usepackage{latexsym}

\usepackage[T1]{fontenc}

\usepackage[utf8]{inputenc}

\usepackage{microtype}

\usepackage{inconsolata}

\usepackage{graphicx}

\usepackage{booktabs, multirow} 
\usepackage{soul}
\usepackage{xcolor,colortbl} 
\usepackage{changepage,threeparttable} 
\usepackage{amsmath}
\usepackage{rotating}
\usepackage{todonotes}

\definecolor{myOrange}{HTML}{ffa30f }
\definecolor{myBlue}{HTML}{2b3cf2}

\title{Toppling the Hierarchy in Byte-level Language Modeling}

\author{Lukas Edman$^{1,2}$\qquad Alexander Fraser$^{1,2,3}$ \vspace{.2cm}\\ 
$^{1}$School of Computation, Information and Technology, TU Munich \\
$^{2}$Munich Center for Machine Learning \\
$^{3}$Munich Data Science Institute \\
\vspace{.1cm} {\tt \small lukas.edman@tum.de}
}

\begin{document}
\maketitle
\begin{abstract}

This work examines recent byte-level models and their failure to perfectly manipulate characters. State-of-the-art byte-level models use a hierarchical structure, starting at the byte level, downsampling to the word level, and then upsampling back to bytes. While this improves training and inference efficiency, we find that the hierarchical design itself limits character-level understanding, with pure byte-level models consistently outperforming hierarchical variants on character manipulation tasks. Ablating transformer layers into attention and feed-forward components further reveals that byte-level attention is the primary mechanism driving this behavior. Together, our results provide an explanation for the character-level failures of hierarchical byte models and establish a clear trade-off between computational efficiency and fine-grained character understanding.



\end{abstract}

\section{Introduction}
Byte-level or character-level models have shown promise in recent years, with several works boasting superior performance over BPE-based subword models on character-level manipulation benchmarks (i.e., CUTE \cite{edman-etal-2024-cute} and the multilingual version EXECUTE \cite{edman-etal-2025-execute}, while keeping up-to-par on other popular benchmarks (i.e., ARC \cite{clark2018arc}, MMLU \cite{hendrycks2021mmlu}, PiQA \cite{bisk2019piqa}, \textit{inter alia}).

However, despite the performance being superior to subword models, it is still far from perfect. The Byte Latent Transformer \cite{pagnoni-etal-2025-byte}, which is pretrained from scratch, only achieved an average of 54.1\% on CUTE, and the more recent Bolmo model \cite{minixhofer2026bolmo}, which is adapted from Olmo, only increased CUTE performance from its equivalent subword-level counterpart by around 6\% (78.6\% vs. 72.9\%). Meanwhile, \citet{edman-etal-2025-execute} found that even BPE-based LLMs can achieve remarkable, near 100\%, few-shot EXECUTE scores in Amharic and other low-resource languages with almost no pre-training data. One reason for this stellar performance may be that these LLMs typically dedicate no BPE merges to the Amharic script and other minority scripts, so with byte fallback, they end up operating entirely on the byte-level, making character-level manipulation easier. 

\begin{figure}[tp!]
    \centering
    \includegraphics[width=\linewidth]{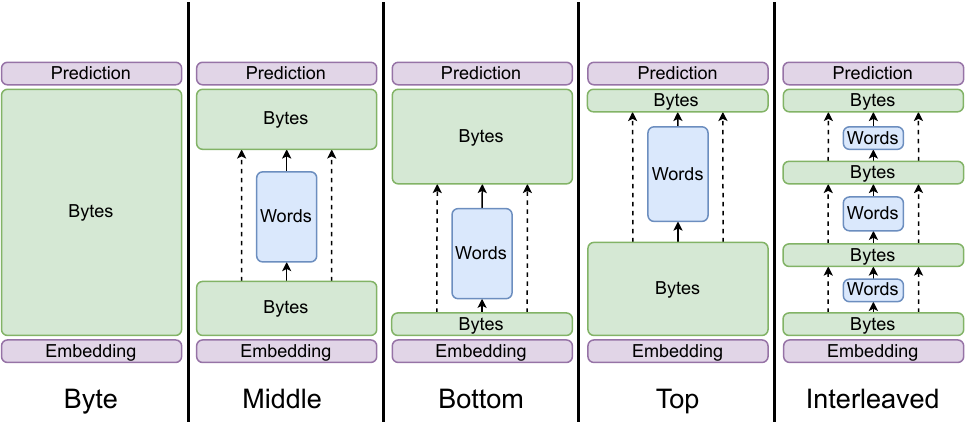}
    \caption{Hierarchies tested in this work, with word-level transformer layers in the \texttt{middle} (a.k.a., hourglass), on the \texttt{bottom}, \texttt{top}, and \texttt{interleaved}. Dashed lines indicate skip connections.}
    \label{fig:hierarchies}
\end{figure}


These findings raise doubts about the dominant hierarchical approach to byte-level modeling, where most computation occurs at the word level through word-level attention and feed-forward layers. Although byte-level attention and feed-forward layers substantially increase training and inference cost, EXECUTE's Amharic results suggest they may be critical for strong character-level manipulation. Understanding their role is therefore key to optimizing the performance--efficiency trade-off.


In this work, we train models from scratch, enabling full control over model architecture and allowing us to answer these questions:
\begin{enumerate}
    \itemsep-3pt
    \item How do hierarchical models fare against pure byte-level models in character-level understanding?
    \item Where is the optimal placement of the byte-level layers within the model?
    \item Which is more critical for character-level understanding,  character-level \textit{attention}, or character-level \textit{feed-forward} modules?
\end{enumerate}
We start by comparing the hierarchical versus the pure byte-level, and explore additional, new strategies for creating a hierarchy. Our contributions are two-fold: 1) We find that a pure byte-level model is best for character-level understanding. 2) If the efficiency gains of a hierarchical model are desired, we present novel hierarchical strategies that better optimize the performance--efficiency trade-off.
We provide code to reproduce our results.\footnote{\url{https://github.com/Leukas/TopplingHierarchy}}



\section{Background}

Hierarchical byte-level models aim to improve the efficiency of pure byte-level models while retaining byte-level input and output. Early approaches used convolutions and RNNs for downsampling and upsampling characters (\citealp{lee-etal-2017-fully,libovicky-etal-2022-dont,edman-etal-2022-subword}, \textit{inter alia}), while recent work uses transformers throughout (\citealp{slagle2024spacebyte,pagnoni-etal-2025-byte,minixhofer2026bolmo}, \textit{inter alia}). Most work focuses on how bytes are grouped into pseudo-word tokens, with strategies including constant, whitespace-, BPE-, and entropy-based sampling. Existing architectures generally follow the same pattern: downsample, process the shorter sequence, then upsample again, including multi-scale approaches such as Hourglass Transformers \cite{nawrot-etal-2022-hierarchical}.

Current state-of-the-art byte-level models include Byte Latent Transformer \cite{pagnoni-etal-2025-byte} and Bolmo \cite{minixhofer2026bolmo}. 
Both of these models have a similar architecture, using a single character-level transformer layer, followed by a large number (16-32) of word-level transformer layers, followed by a smaller number (4-9) for character-level decoder layers. These models perform better similarly to subword models on traditional benchmarks, while also better on character-level understanding tasks such as CUTE. 

Character understanding is commonly evaluated with CUTE \cite{edman-etal-2024-cute}, which tests simple character manipulations such as inserting random characters into words. Its multilingual extension, EXECUTE \cite{edman-etal-2025-execute}, finds that models often perform better on low-resource scripts, likely because these scripts are processed more directly at the byte level and are less constrained by word-level biases.

\begin{table}[!htp]\centering \scriptsize
\begin{tabular}{ll}\toprule
Model &Notation \\\midrule

Byte & $(C_aC_f)^{12} $ \\[2pt]
Top$_6$ & $(C_aC_f)^5\,(W_aW_f)^6\,(C_aC_f)^1$ \\[2pt]
Mid$_6$ & $(C_aC_f)^3\,(W_aW_f)^6\,(C_aC_f)^3$ \\[2pt]
Bot$_6$ & $(C_aC_f)^1\,(W_aW_f)^6\,(C_aC_f)^5$ \\[2pt]
Intl$_6$ & $(C_aC_f\,W_aW_f)^3\,(W_aW_f\,C_aC_f)^3$ \\[2pt]
Mid$_8$ & $(C_aC_f)^2\,(W_aW_f)^8\,(C_aC_f)^2$ \\[2pt]
Mid$_{10}$ & $(C_aC_f)^1\,(W_aW_f)^{10}\,(C_aC_f)^1$ \\[2pt]
Mid$_{10}$ CW & $(C_aC_f)^1\,(C_aW_f)^{10}\,(C_aC_f)^1$ \\[2pt]
Mid$_{10}$ WC & $(C_aC_f)^1\,(W_aC_f)^{10}\,(C_aC_f)^1$ \\

\bottomrule
\end{tabular}
\caption{12-layer models tested and their notation. The number in the model name refers to the number of word-level layers.}\label{tab:models_note}
\end{table}


\begin{table*}[!htp]\centering\scriptsize
\begin{tabular}{lrrrrrrrrrrr}\toprule
&\multicolumn{3}{c}{CUTE $\uparrow$} &\multicolumn{4}{c}{General $\uparrow$} &\multicolumn{2}{c}{Speed $\uparrow$} & Mem $\downarrow$ \\\cmidrule(lr){2-4}\cmidrule(lr){5-8}\cmidrule(lr){9-10}\cmidrule(lr){11-11}
&Pure &Diluted &Avg &ARC-E &Lambada &PIQA &Avg &Train &Infer &Train \\\midrule
Byte &\textbf{94.0} &\textbf{31.2} &\textbf{62.6} &42.0 &\textbf{19.8} &61.7 &41.2 &1.00 &1.00 &1.00 \\
Bot$_6$ &78.5 &23.0 &50.8 &37.9 &15.9 &59.2 &37.7 &1.48 &1.40 &0.65 \\
Top$_6$ &16.6 &27.4 &22.0 &39.6 &19.1 &62.7 &40.5 &1.48 &1.40 &0.65 \\
Intl$_6$ &10.2 &15.2 &12.7 &40.3 &19.1 &62.1 &40.5 &1.48 &1.40 &0.65 \\
Mid$_6$ &37.8 &15.1 &26.4 &37.3 &17.4 &61.9 &38.9 &1.48 &1.40 &0.65 \\
Mid$_8$ &27.1 &14.0 &20.5 &39.8 &18.8 &\textbf{62.8} &40.5 &1.76 &1.63 &0.54 \\
Mid$_{10}$ &0.0 &0.2 &0.1 &39.9 &18.4 &61.2 &39.8 &2.17 &1.93 &\textbf{0.42} \\
Mid$_{10}$ CW &64.0 &23.9 &44.0 &40.2 &16.8 &62.1 &39.7 &1.31 &1.28 &0.58 \\
Mid$_{10}$ WC &1.4 &3.8 &2.6 &39.7 &16.8 &61.0 &39.2 &1.43 &1.35 &0.84 \\
Subword & 35.5 & 23.1 &	29.3 &\textbf{45.7} &	15.8 &	\textbf{65.8} &	\textbf{42.4} & \textbf{2.21} & \textbf{2.81} & 0.92 \\

\bottomrule
\end{tabular}
\caption{Results on 12-layer models, on \texttt{Pure} and \texttt{Diluted} CUTE data, as well as standard benchmarks. We represent speed and memory as a ratio in relation to \texttt{Byte}.}\label{tab:main_results}
\end{table*}

\section{Methodology}
Given that our interest is in determining the architectural conditions for good character-level understanding, we develop a notation for the larger class of hierarchies we investigate. 

\subsection{Hierarchical Notation}

Prior work on hierarchical byte-level models typically places word-level transformer layers either centrally, with similar numbers of character-level layers on both sides (\texttt{Middle}), or later in the model, with more character-level layers preceding them (\texttt{Bottom}). We also study a \texttt{Top} strategy, where most character-level layers come first, and an \texttt{Interleaved} strategy, where character- and word-level layers alternate (Figure \ref{fig:hierarchies}).

Both attention and feed-forward modules can operate at either the byte or word level. Byte-level modules process all tokens, while word-level modules act only on masked subsets, with updates added back to the residual stream. Following \citet{slagle2024spacebyte}, we sample using whitespace delimiters, which works ``surprisingly well in practice for English texts''.

We denote architectures using four module types: character-level attention ($C_a$), character-level feed-forward ($C_f$), word-level attention ($W_a$), and word-level feed-forward ($W_f$). Repeated patterns use exponents, e.g., $(C_aC_f)^{12}$ for a standard 12-layer byte-level Transformer. Concatenation proceeds left-to-right, so Bolmo's hierarchy is written as $(C_aC_f)^1\,(W_aW_f)^{N}\,(C_aC_f)^4$, with $N=\{16,32\}$. Tables \ref{tab:models_note} and \ref{tab:hr_models_note} list the 12- and 24-layer models tested, enabling control over word-layer placement, count, and module type. All models begin and end with fully character-level layers to ensure full token accessibility and avoid dead ends in the computation graph.




\subsection{Experimental Setup}

We evaluate two model scales: 12-layer models with hidden size 768 and 24-layer models with hidden size 1024, corresponding to roughly 100M and 400M parameters with a byte-level vocabulary. Following the scaling laws of \citet{qiu2026hyperparameter}, which suggest 9.6  GPT-2 tokens per parameter, we treat ``GPT-2 tokens'' as roughly equivalent to words and train on 1B- and 4B-word subsets of Nemotron-ClimbMix \cite{diao2025climb} for the two scales.

The 1B-word experiments compare hierarchies and module types, while the larger-scale 4B experiments test whether the number of character-level layers should remain fixed or scale with model depth. Prior work differs on this choice: \citet{slagle2024spacebyte} scaled character layers proportionally, whereas \citet{minixhofer2026bolmo} kept them constant. We investigate which better supports character understanding.

We further continue training on a \texttt{Diluted} dataset combining additional ClimbMix data with 1\% CUTE training data from \citet{minixhofer2026bolmo}, using 500 examples per subtask.\footnote{We omit CUTE's \texttt{orth} and \texttt{sem} subtasks as training data is not available.} This setup tests whether models can balance character-level and general language capabilities, similar to standard fine-tuning approaches.

We also train on a \texttt{Pure} dataset containing only CUTE data, again with 500 examples per subtask, to estimate the upper bound of character-understanding performance.



Training hyperparameters are reported in Appendix \ref{app:hyperparams}. The longest run (a byte-level model on 4B words) required about five days on a single H100 GPU.

Evaluation focuses on CUTE, however we also report an average of standard NLP benchmarks: Lambada \cite{paperno-etal-2016-lambada}, PIQA \cite{bisk2019piqa}, and ARC-Easy \cite{clark2018arc}. This is to assess whether architectural changes impact general NLP ability.\footnote{We omit more knowledge-intensive benchmarks such as MMLU \cite{hendrycks2021mmlu} due to limited training scale.}

\section{Results}
We show our results on the 12-layer models in Table \ref{tab:main_results}. We also include a subword model for reference, which uses the Olmo 2 vocabulary. In addition to CUTE and general benchmark scores, we show the training and inference speed, as well as training GPU memory usage, in relation to the \texttt{Byte} model.\footnote{Inference memory usage is practically the same for all models. The efficiency metrics of \textit{Subword} are with embeddings being on the GPU. }
Overall, we see a significantly stronger performance for \texttt{Byte} on both the pure and diluted CUTE, while the general benchmarks have no clear winner, only mildly favoring \texttt{Byte} and \texttt{Subword}. 
We now break down the results according to our research questions.

\paragraph{Layer distribution}
Comparing our different layering strategies (i.e., \texttt{Bot, Mid, Top, Intl}), the results indicate a preference for more character-level layers in the latter half of the transformer, with stronger performances from \texttt{Bot} and \texttt{Mid} on \texttt{Pure} CUTE. This tracks with the general understanding that these latter layers perform task-specific operations, where the ability to operate on the character-level would be more useful for CUTE. The diluted CUTE results appear more evenly balanced, suggesting the concentration of character-level layers on the top is only better when the task is focused.

\paragraph{Number of word-level layers}
We increase the number of word-level layers from \texttt{Mid$_6$} up to \texttt{Mid$_{10}$}. As the number of word-level layers increases, the performance drops. This is not surprising, as fewer layers tends towards the behavior of a subword model, but with the added difficulty of a small, byte-level vocabulary. 

\paragraph{Attention vs. Feed-forward}
The scores from \texttt{Byte}, \texttt{Mid$_{10}$}, \texttt{Mid$_{10}$ CW}, and \texttt{Mid$_{10}$ WC} paint a clear picture: attention is more important for character-level understanding. The two models with fully character-level attention, \texttt{Byte} and \texttt{Mid$_{10}$ CW}, substantially outperform those with mostly word-level attention. 

\begin{table}[!tp]\centering\scriptsize
\begin{tabular}{lrrrrrrr}\toprule
&\multicolumn{3}{c}{CUTE $\uparrow$} &\multicolumn{2}{c}{Speed $\uparrow$} & Mem $\downarrow$ \\\cmidrule(lr){2-4}\cmidrule(lr){5-6} \cmidrule(lr){7-7}
&Pure &Dilu. &Avg &Train &Infer & Train \\\midrule
Byte &\textbf{97.5} &\textbf{55.8} &\textbf{76.6} &1.00 &1.00 &1.00 \\
Top$_{12}$ &91.7 &51.2 &71.5 &1.51 &1.45 &0.65 \\
Intl$_{12}$ &95.9 &43.1 &69.5 &1.51 &1.45 &0.65 \\
Intl$_{18}$ &91.8 &44.3 &68.1 &2.04 &1.87 &\textbf{0.47} \\
Intl$_{\text{hyb}}$ &92.6 &43.0 &67.8 &1.72 &1.62 &0.52 \\
Mid$_{12}$ &94.3 &40.7 &67.5 &1.51 &1.45 &0.65 \\
Mid$_{\text{hyb}}$ &96.0 &38.5 &67.2 &1.72 &1.62 &0.52 \\
Bot$_{\text{hyb}}$ &93.0 &40.4 &66.7 &1.72 &1.62 &0.52 \\
Top$_{18}$ &91.4 &38.7 &65.0 &2.04 &1.87 &\textbf{0.47} \\
Bot$_{12}$ &96.3 &30.5 &63.4 &1.51 &1.45 &0.65 \\
Top$_{\text{hyb}}$ &87.5 &38.2 &62.9 &1.72 &1.62 &0.52 \\
Mid$_{18}$ &91.6 &26.4 &59.0 &2.04 &1.87 &\textbf{0.47} \\
Bot$_{18}$ &92.0 &17.8 &54.9 &2.04 &1.87 &\textbf{0.47} \\
Subword & 46.4 & 33.1 & 39.7 & \textbf{3.21} & \textbf{3.62} & 0.52 \\

\bottomrule
\end{tabular}
\caption{Results on 24-layer models, sorted by average CUTE performance. We represent speed and memory as a ratio in relation to \texttt{Byte}.}\label{tab:hr_results}
\end{table}

\paragraph{Scaling Up}

In Table \ref{tab:hr_results}, we present higher-resourced models ordered by average CUTE performance. Following the positive results of the $C_aW_f$ layer, we also evaluate hybrid architectures, including models with 6 $C_aW_f$ layers, 12 $W_aW_f$ layers, and 6 $C_aC_f$ layers (see Table \ref{tab:hr_models_note} for details). We omit non-CUTE results for brevity; full results are in Appendix \ref{app:full_results}.

Overall, the \texttt{Byte} model performs best by a 5\% margin, showing stronger character understanding in both focused and general settings. We also note that the \texttt{Byte} model from the 1B setting, having roughly 8$\times$ less compute, performs better than \texttt{Mid$_{18}$} and \texttt{Bot$_{18}$} (62.6 vs. 59.0 and 54.9).
In general, more character-level layers improve performance. On \texttt{Pure}, \texttt{Mid} and \texttt{Bot} perform strongly, but are lacking on \texttt{Diluted}, while \texttt{Top$_{12}$} and \texttt{Interleaved} models are more robust in that setting. Hybrid models (apart from \texttt{Top}) show similar performance to those with 12 $C_aC_f$ layers, indicating $C_aW_f$ may be a cost-saving alternative to $C_aC_f$.


Despite training on only 4B words, nearly all models exceed 90\% on \texttt{Pure}, substantially outperforming Bolmo despite its far larger training budget (49B adaptation tokens, 4T+ pretraining tokens, including 21M CUTE tokens). However, CUTE represents only 0.04\% of its adaptation data, which explains its weaker retention of character-level knowledge.

Scaling trends show overall improvement, but architecture still matters. For example, \texttt{Bot$_{18}$}, which most closely resembles Bolmo and Byte Latent Transformer, improves only 4.1 points over \texttt{Bot$_6$} when scaled 4$\times$ in model and data size. In contrast, \texttt{Intl$_{18}$} improves 55.4 points, suggesting better scaling behavior and faster convergence toward the performance of \texttt{Byte}.

Overall, \texttt{Byte} models achieve the best performance, while \texttt{Interleaved} models offer a strong trade-off, enabling up to 2$\times$ faster training and reduced memory usage (less than half the RAM).

\section{Conclusion}
State-of-the-art byte-level models boast a better performance on character-level understanding tasks such as CUTE, however we have shown that achieving much better performance can be done \textit{without} a hierarchical structure that downsamples to the word-level. We test several types of hierarchical structures, and all are outperformed by the fully byte-level models. However, due to their speed and memory efficiency, hierarchical models still may be desirable. In this case, we suggest an interleaved strategy, with alternating character-level and word-level layers. 

There are still several open questions with byte-level models left unanswered. Concerning adaptation from a pretrained subword model (e.g., Bolmo from Olmo), an interleaved strategy may require a novel approach. Speeding up fully byte-level models would also be highly desirable.

\section*{Limitations}

Our experiments are limited to English. We do not expect the results will differ in Latin-scripted languages, however languages such as Chinese may exhibit different trends due to the larger character set and shorter words. The whitespace sampling strategy we borrow from \citet{slagle2024spacebyte} would not be suitable for some languages.

We also cannot exhaustively search all possible module configurations, so we limit the models tested to always alternating between attention and feedforward modules, as this is an established property of the Transformer architecture. However, even within this self-imposed limitation, we cannot test every possible configuration. We notably did not test models with a higher ratio of byte-level layers to word-level layers (e.g., \texttt{Mid$_2$} or \texttt{Mid$_4$}). We expect these models to perform more and more similarly to the fully byte-level model, but they will consequently also have an increasingly similar speed and memory usage. As we have not seen any such models in prior research, we limit our study to models with an equal or higher word-to-character layer ratio.

Our experiments are on the 1B to 4B word scale, so we cannot definitively draw conclusions on very low resource or very high resource settings. Our trends indicate possible convergence of some of the hierarchical models to the byte-level performance, however it is not clear whether this indeed happens. We nevertheless show the capacity for better performance via our experiments on the \texttt{Pure} dataset.

\section*{Acknowledgments}
The authors gratefully acknowledge the scientific support and HPC resources provided by the Erlangen National High Performance Computing Center (NHR@FAU) of the Friedrich-Alexander-Universität Erlangen-Nürnberg (FAU) under the NHR project b279bb. NHR funding is provided by federal and Bavarian state authorities. NHR@FAU hardware is partially funded by the German Research Foundation (DFG) – 440719683.
This work was co-funded by the European Union (ERC, EPICAL, 101141712). Views and opinions expressed are however those of the author(s) only and do not necessarily reflect those of the European Union or the European Research Council. Neither the European Union nor the granting authority can be held responsible for them.

\bibliography{custom}

\appendix

\section{24-layer models} \label{app:hr_models_note}
We show the notation for 24-layer models in Table \ref{tab:hr_models_note}. The models follow a similar recipe to the 12-layer models (in Table \ref{tab:models_note}), but adjusted to the 24-layer scale. Models denoted $X_{12}$ have 12 word-level layers, which is a 1:1 word/byte layer ratio. Models denoted $X_{18}$ have 18 word-level layers, with a 3:1 word/byte layer ratio, keeping the 6 byte layers constant from the 12-layer model experiments. Hybrid models are in between: 6 fully byte layers, 6 hybrid $C_aW_f$ layers, and 12 fully word layers.
\label{sec:appendix}

\begin{table*}[!htp]\centering \small
\begin{tabular}{ll}\toprule
Model &Notation \\\midrule
Byte & $(C_aC_f)^{24} $ \\[2pt]
Top$_{12}$ & $(C_aC_f)^{11}\,(W_aW_f)^{12}\,(C_aC_f)^1$ \\[2pt]
Top$_{18}$ & $(C_aC_f)^5\,(W_aW_f)^{18}\,(C_aC_f)^1$ \\[2pt]
Top$_{\text{hyb}}$ & $(C_aC_f)^5\,(C_aW_f)^{6}\,(W_aW_f)^{12}\,(C_aC_f)^1$ \\[2pt]
Mid$_{12}$ & $(C_aC_f)^6\,(W_aW_f)^{12}\,(C_aC_f)^6$ \\[2pt]
Mid$_{18}$ & $(C_aC_f)^3\,(W_aW_f)^{18}\,(C_aC_f)^3$ \\[2pt]
Mid$_{\text{hyb}}$ & $(C_aC_f)^3\,(C_aW_f)^{3}\,(W_aW_f)^{12}\,(C_aW_f)^{3}\,(C_aC_f)^3$ \\[2pt]
Bot$_{12}$ & $(C_aC_f)^1\,(W_aW_f)^{12}\,(C_aC_f)^{11}$ \\[2pt]
Bot$_{18}$ & $(C_aC_f)^1\,(W_aW_f)^{18}\,(C_aC_f)^5$ \\[2pt]
Bot$_{\text{hyb}}$ & $(C_aC_f)^1\,(W_aW_f)^{12}\,(C_aW_f)^{6}\,(C_aC_f)^5$ \\[2pt]
Intl$_{12}$ & $(C_aC_f\,W_aW_f)^6\,(W_aW_f\,C_aC_f)^6$ \\[2pt]
Intl$_{18}$ & $(C_aC_f\,(W_aW_f)^3)^2\,((W_aW_f)^3\,C_aC_f)^2$ \\[2pt]
Intl$_{\text{hyb}}$ & $(C_aC_f\,W_aW_f\,C_aW_f\,W_aW_f)^2\,(W_aW_f\,C_aW_f\,W_aW_f\,C_aC_f)^2$ \\[2pt]
\bottomrule
\end{tabular}
\caption{24-layer models tested and their notation. The number in the model name refers to the number of word-level layers.}\label{tab:hr_models_note}
\end{table*}

\section{Hyperparameters} \label{app:hyperparams}
We show the hyperparameters used in our experiments in Table \ref{tab:hyperparams}. We vary the batch size in our ClimbMix experiments to keep the number of iterations roughly equal. In initial testing, we used a learning rate of $4e-4$, following Byte Latent Transformer, but scaled down to $2e-4$ due to the smaller scales requiring smaller batch sizes.
\begin{table*}[!htp]\centering\small
\begin{tabular}{lrrrrr}\toprule
Dataset &ClimbMix 1B &ClimbMix 4B &Diluted &Pure \\\midrule
Batch size &32 &128 &32 &8 \\
Sequence length &2048 &2048 &2048 &2048 \\
Epochs &1 &1 &5 &5 \\
LR &2e-4 &2e-4 &1e-4 &2e-5 \\
Warmup steps &2000 &2000 &2000 &50 \\
Optimizer &AdamW &AdamW &AdamW &AdamW \\
Betas &(.9, .98) &(.9, .98) &(.9, .98) &(.9, .999) \\
Weight decay &0.01 &0.01 &0.01 &0.01 \\
Grad norm clip &1 &1 &1 &1 \\
Precision &bf16 &bf16 &bf16 &bf16 \\
\bottomrule
\end{tabular}
\caption{Hyperparameters used for our experiments.}\label{tab:hyperparams}

\end{table*}

\section{Complete Results} \label{app:full_results}
All of the results obtained in our experiments are presented in Table \ref{tab:full_results}. Concerning the CUTE subtasks, we also split them based on character- versus word-level tasks, but found no meaningful difference in the trends. We suspect this is because all of the models tested are byte-level, so the word-level tasks of CUTE can be thought of as several sequential byte-level operations. 

\begin{sidewaystable*}[!htp]\centering \tiny

\begin{tabular}{lrrrrrrrrrrrrrrrrrr}\toprule
Size &Dataset & Model &Spell &Spell Inv &Cont C &Cont W &Ins C &Ins W &Del C &Del W &Sub C &Sub W &Swap C &Swap W &ARC-E &Lambada &PIQA \\\midrule
1B &Diluted &\texttt{Byte} &70.9 &53.6 &64.8 &64.6 &12.4 &7.2 &28.0 &24.5 &21.8 &17.7 &6.4 &2.1 &42.0 &19.8 &61.7 \\
1B &Diluted &\texttt{Bot$_6$} &70.3 &50.6 &57.9 &52.4 &2.1 &3.7 &6.7 &6.8 &10.5 &6.3 &6.3 &2.6 &37.9 &15.9 &59.2 \\
1B &Diluted &\texttt{Top$_6$} &36.7 &53.3 &51.2 &57.5 &6.0 &13.0 &33.8 &16.2 &36.8 &16.5 &5.1 &2.4 &39.6 &19.1 &62.7 \\
1B &Diluted &\texttt{Intl$_6$} &20.4 &10.5 &38.9 &51.3 &0.9 &8.8 &10.4 &16.2 &10.4 &13.1 &0.5 &0.9 &37.3 &17.4 &61.9 \\
1B &Diluted &\texttt{Mid$_6$} &5.6 &17.4 &58.0 &62.1 &3.2 &1.4 &6.4 &2.3 &14.9 &4.0 &5.3 &0.4 &40.3 &19.1 &62.1 \\
1B &Diluted &\texttt{Mid$_8$} &0.2 &33.2 &49.3 &53.1 &0.9 &1.4 &3.1 &5.3 &15.3 &3.9 &2.4 &0.0 &39.8 &18.8 &62.8 \\
1B &Diluted &\texttt{Mid$_10$} &0.0 &0.3 &0.0 &0.0 &0.0 &0.0 &0.0 &1.1 &0.7 &0.3 &0.0 &0.0 &39.9 &18.4 &61.2 \\
1B &Diluted &\texttt{Mid$_{10}$ CW} &54.2 &67.7 &56.2 &65.5 &1.8 &7.5 &7.3 &8.2 &9.4 &8.5 &0.4 &0.5 &40.2 &16.8 &62.1 \\
1B &Diluted &\texttt{Mid$_{10}$ WC} &0.0 &9.6 &5.2 &21.5 &0.1 &0.4 &0.9 &4.6 &2.0 &1.1 &0.1 &0.0 &39.7 &16.8 &61.0 \\
1B &Pure &\texttt{Byte} &99.2 &99.5 &100.0 &99.4 &91.4 &90.8 &97.7 &90.2 &98.5 &93.5 &95.4 &71.9 &- &- &- \\
1B &Pure &\texttt{Bot$_6$} &99.6 &99.1 &69.9 &81.8 &61.4 &80.1 &91.1 &78.9 &92.4 &71.1 &81.9 &35.3 &- &- &- \\
1B &Pure &\texttt{Top$_6$} &0.1 &40.9 &24.4 &19.3 &8.1 &0.1 &39.2 &0.1 &55.9 &0.0 &10.6 &0.0 &- &- &- \\
1B &Pure &\texttt{Intl$_6$} &40.7 &88.6 &5.2 &1.5 &70.2 &0.0 &88.4 &4.7 &85.1 &1.0 &67.8 &0.0 &- &- &- \\
1B &Pure &\texttt{Mid$_6$} &0.0 &31.7 &0.0 &1.6 &14.6 &0.0 &21.8 &1.2 &46.1 &0.1 &4.8 &0.0 &- &- &- \\
1B &Pure &\texttt{Mid$_8$} &8.6 &58.4 &0.0 &0.0 &76.1 &0.0 &68.7 &3.0 &71.5 &1.9 &36.7 &0.0 &- &- &- \\
1B &Pure &\texttt{Mid$_10$} &0.0 &0.0 &0.0 &0.0 &0.0 &0.0 &0.0 &0.0 &0.0 &0.0 &0.0 &0.0 &- &- &- \\
1B &Pure &\texttt{Mid$_{10}$ CW} &98.6 &99.0 &69.1 &79.5 &29.5 &71.5 &76.2 &65.1 &76.7 &53.7 &31.7 &17.6 &- &- &- \\
1B &Pure &\texttt{Mid$_{10}$ WC} &0.0 &7.1 &0.0 &0.0 &2.3 &0.1 &0.9 &0.0 &6.9 &0.0 &0.1 &0.0 &- &- &- \\
4B &Diluted &\texttt{Byte} &73.3 &86.4 &65.6 &69.6 &33.6 &58.3 &59.1 &53.4 &60.8 &62.1 &25.3 &21.7 &46.4 &25.4 &65.6 \\
4B &Diluted &\texttt{Bot$_{12}$} &33.0 &11.1 &54.1 &66.3 &9.9 &29.5 &46.3 &27.0 &21.9 &21.9 &29.2 &15.3 &45.7 &23.0 &65.1 \\
4B &Diluted &\texttt{Bot$_{18}$} &10.8 &14.1 &46.3 &54.4 &0.2 &6.4 &47.9 &12.5 &6.7 &4.6 &8.3 &2.0 &48.6 &23.3 &66.1 \\
4B &Diluted &\texttt{Bot$_{\text{hyb}}$} &30.4 &63.5 &54.0 &75.4 &7.2 &19.1 &62.6 &29.2 &51.9 &25.0 &54.2 &12.3 &48.6 &21.7 &65.7 \\
4B &Diluted &\texttt{Mid$_{12}$} &77.0 &75.4 &20.4 &69.5 &20.0 &31.3 &46.7 &24.3 &57.6 &35.0 &22.2 &8.5 &48.2 &25.3 &65.1 \\
4B &Diluted &\texttt{Mid$_{18}$} &19.0 &53.3 &59.3 &74.9 &10.6 &7.6 &18.4 &6.3 &31.4 &26.7 &8.3 &1.3 &49.6 &22.3 &66.8 \\
4B &Diluted &\texttt{Mid$_{\text{hyb}}$} &66.6 &79.5 &63.3 &63.9 &12.3 &13.5 &29.8 &28.9 &43.5 &41.6 &9.1 &9.5 &48.8 &23.7 &66.8 \\
4B &Diluted &\texttt{Intl$_{12}$} &46.9 &86.8 &65.4 &78.6 &22.8 &55.3 &55.2 &44.0 &63.6 &55.9 &19.2 &20.6 &45.8 &20.9 &65.6 \\
4B &Diluted &\texttt{Intl$_{18}$} &54.9 &77.5 &60.8 &59.7 &12.3 &44.2 &28.0 &31.0 &28.5 &46.0 &2.1 &0.0 &48.1 &23.0 &64.0 \\
4B &Diluted &\texttt{Intl$_{\text{hyb}}$} &4.5 &66.1 &65.6 &59.8 &9.0 &48.4 &39.0 &38.4 &46.0 &50.8 &16.5 &14.7 &47.1 &23.2 &66.6 \\
4B &Diluted &\texttt{Top$_{12}$} &63.3 &61.5 &67.9 &66.6 &22.3 &45.6 &42.1 &28.0 &41.7 &49.9 &16.2 &12.3 &46.3 &23.4 &64.4 \\
4B &Diluted &\texttt{Top$_{18}$} &78.4 &80.9 &57.7 &65.4 &9.8 &37.7 &40.2 &26.3 &61.6 &40.6 &20.9 &12.3 &46.1 &22.5 &65.4 \\
4B &Diluted &\texttt{Top$_{\text{hyb}}$} &77.3 &61.7 &64.2 &71.0 &7.9 &38.5 &41.5 &13.0 &72.1 &20.1 &42.3 &0.0 &48.4 &21.7 &64.5 \\
4B &Pure &\texttt{Byte} &99.9 &99.7 &99.8 &99.7 &95.0 &96.6 &99.4 &94.4 &98.6 &97.4 &97.1 &92.8 &- &- &- \\
4B &Pure &\texttt{Bot$_{12}$} &100.0 &99.9 &99.9 &99.9 &96.2 &95.5 &98.2 &90.8 &99.2 &94.5 &98.4 &82.5 &- &- &- \\
4B &Pure &\texttt{Bot$_{18}$} &100.0 &99.5 &72.2 &99.7 &90.4 &96.6 &95.7 &90.3 &94.9 &95.4 &84.4 &84.9 &- &- &- \\
4B &Pure &\texttt{Bot$_{\text{hyb}}$} &99.8 &100.0 &70.3 &99.5 &90.4 &95.7 &95.2 &90.8 &98.5 &95.0 &96.6 &84.8 &- &- &- \\
4B &Pure &\texttt{Mid$_{12}$} &99.9 &99.7 &99.7 &100.0 &92.4 &95.6 &96.8 &91.1 &94.8 &96.9 &77.2 &87.3 &- &- &- \\
4B &Pure &\texttt{Mid$_{18}$} &99.2 &99.7 &71.3 &99.9 &92.3 &95.7 &94.2 &89.6 &92.7 &96.7 &79.7 &87.8 &- &- &- \\
4B &Pure &\texttt{Mid$_{\text{hyb}}$} &99.1 &99.6 &99.4 &99.4 &93.6 &95.8 &97.7 &92.0 &98.3 &97.4 &93.1 &86.5 &- &- &- \\
4B &Pure &\texttt{Intl$_{12}$} &99.8 &99.6 &99.4 &99.5 &93.9 &95.2 &98.2 &91.4 &97.1 &97.8 &91.0 &88.3 &- &- &- \\
4B &Pure &\texttt{Intl$_{18}$} &99.2 &98.8 &99.7 &99.7 &87.4 &94.5 &94.6 &89.9 &91.7 &95.6 &67.5 &83.1 &- &- &- \\
4B &Pure &\texttt{Intl$_{\text{hyb}}$} &99.5 &99.1 &77.8 &99.2 &89.4 &93.6 &91.5 &90.2 &97.7 &95.8 &95.1 &81.9 &- &- &- \\
4B &Pure &\texttt{Top$_{12}$} &99.6 &99.4 &98.9 &99.9 &86.9 &91.5 &92.9 &86.4 &92.5 &94.8 &79.0 &78.8 &- &- &- \\
4B &Pure &\texttt{Top$_{18}$} &98.6 &99.2 &99.8 &99.9 &78.8 &92.3 &90.1 &88.6 &93.2 &95.5 &77.4 &83.1 &- &- &- \\
4B &Pure &\texttt{Top$_{\text{hyb}}$} &99.2 &99.7 &69.6 &99.8 &71.3 &91.6 &85.1 &87.9 &91.5 &93.3 &80.0 &80.7 &- &- &- \\
\bottomrule
\end{tabular}
\caption{Complete results.}\label{tab:full_results}
\end{sidewaystable*}

\end{document}